\documentclass[twoside,11pt]{article}
\usepackage{jmlr2e}
\usepackage{amsmath}
\usepackage{booktabs}
\usepackage{float}

\title{Building Efficient Lightweight CNN Models}

\author{\name Isong Nathan \email  isongneee2020@futa.edu.ng \\
        \addr Federal University of Technology, Akure}

\begin{document}

\begin{abstract}
Convolutional Neural Networks (CNNs) are pivotal in image classification tasks due to their robust feature extraction capabilities. However, their high computational and memory requirements pose challenges for deployment in resource-constrained environments. This paper introduces a methodology to construct lightweight CNNs while maintaining competitive accuracy. The approach integrates two stages of training; dual-input-output model and transfer learning with progressive unfreezing. The dual-input-output model train on original and augmented datasets, enhancing robustness. Progressive unfreezing is applied to the unified model to optimize pre-learned features during fine-tuning, enabling faster convergence and improved model accuracy.

The methodology was evaluated on three benchmark datasetshandwritten digit MNIST, fashion MNIST, and CIFAR-10. The proposed model achieved a state-of-the-art accuracy of 99\% on the handwritten digit MNIST and 89\% on fashion MNIST, with only 14,862 parameters and a model size of 0.17 MB. While performance on CIFAR-10 was comparatively lower (65\% with less than 20,00 parameters), the results highlight the scalability of this method. The final model demonstrated fast inference times and low latency, making it suitable for real-time applications.

Future directions include exploring advanced augmentation techniques, improving architectural scalability for complex datasets, and extending the methodology to tasks beyond classification. This research underscores the potential for creating efficient, scalable, and task-specific CNNs for diverse applications.
\end{abstract}


\ShortHeadings{CNN Complexity Reduction}{Nathan}

\editor{}

\maketitle

\section{Introduction}
Convolutional Neural Networks (CNNs) are a cornerstone of image classification tasks, enabling robust feature extraction from visual data, as demonstrated in various works like~\cite{szegedy2015rethinking, huang2018densely}. Despite their success, CNNs are often characterized by high computational demands and significant memory requirements, making their training difficult or requiring costly devices and time for training. Moreover, their deployment is equally challenging in resource-constrained environments such as mobile devices and embedded systems due to their sizes and latencies~\citep{liu2022lightweight}. This limitation has driven the development of several methods such as pruning~\citep{han2015learning}, quantization \citep{jacob2018quantization}, and and filter decomposition \citep{denton2014filterdecomposition}, aiming to reduce CNN models complexity while trying to maintain competitive accuracy.

These traditional techniques for reducing model complexity, primarily optimize pre-trained models. While effective, these methods often require additional retraining cycles or specialized hardware support, becoming time consuming and technically demanding~\citep{lee2018retraining}.

This paper explores a combination of methodologies for constructing lightweight CNNs designed from the ground up to be efficient and accurate. The proposed approach integrates:
\begin{itemize}
    \item \textbf{Dual-Input-Output Model:} A model containing two submodels trained independently but simultaneously—one on the original dataset and the other on an augmented version. On the later stage, the submodels outputs are concatenated, to improve robustness and reduce overfitting through complementation.
    \item \textbf{Transfer Learning with Progressive Unfreezing:} Transfer learning, as defined by~\cite{pan2010transfer}, enables leveraging knowledge gained in one domain to improve learning efficiency in another related domain. In this methodology, a unified model is created by combining the submodels’ weights, followed by some additional layers and fine-tuning using progressive unfreezing. Progressive unfreezing, as proposed by~\cite{howard2018universal}, involves gradually unfreezing and fine-tuning the layers of the model, starting from the last layer, optimizing pre-learned features to enable faster convergence with fewer parameters.
\end{itemize}

\begin{figure}
    \centering
    \includegraphics[width=1\linewidth]{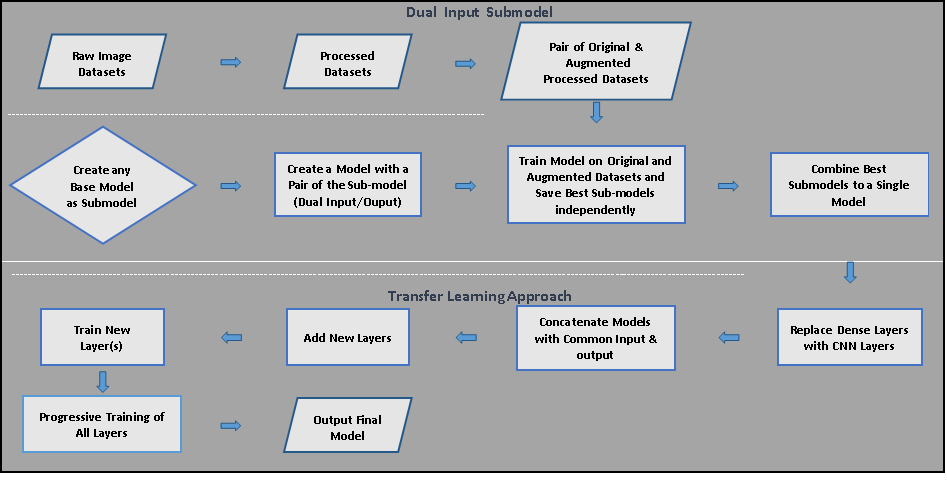}
    \caption{Flow Chart showing the methodology step.}
    \label{fig:flowchart}
\end{figure}

The flowchart for the methodology is shown in figure~\ref{fig:flowchart}

The methodology is applied to three benchmark datasets—handwritten digit MNIST, fashion MNIST, and CIFAR-10-to show the power and the limitation of the one relative simple model as the dataset complexity increases. A careful design of models using this method can achieve a state-of-the-art accuracy when specially designed to fit the dataset. For instance, the approach achieves an accuracy of \texttt{99\%} on the handwritten digit MNIST dataset and \texttt{89\%} on the fashion MNIST dataset evaluated on their test sets (10,000 instances), with a parameter count of only 14,862 and a model size of 0.17 MB.

This research underscores the potential for developing lightweight CNN architectures that are efficient, scalable, and suitable for resource-constrained applications, not only that, but to develop models that just fit into a task rather than using a complex model for a less complex task which comes with its own overhead; the case of not using a machete to cut one's finger nails. However, refining architectures to balance simplicity and performance, and extending the approach to tasks beyond classification, remain critical avenues for future work.

\section{Related Work}
The technique of pruning~\citep{han2015learning} removes unnecessary weights and connections from CNNs, thus improving both speed and memory usage. Despite its benefits, pruning is computationally expensive, requiring long iterative pruning and retraining cycles, which makes it time consuming and unreliable for faster development applications. The pruning process may also fail to provide substantial improvements in runtime speed, especially for deeper networks, and it can be challenging to implement directly in complex architectures.

Quantization~\citep{jacob2018quantization} offers a way to compress CNNs by reducing the precision of model weights and activations, thus lowering memory and computational requirements. While effective in reducing the model size, quantization does not address structural inefficiencies and may introduce errors, particularly in models that rely on higher precision computations. The performance of quantized models can be degraded, particularly in tasks requiring high-accuracy predictions.

The 1D-FALCON scheme~\citep{maji2018complexity_reduction} proposes a filter decomposition method coupled with the Toom-Cook algorithm to accelerate convolution operations in CNNs. This technique significantly reduces the computational load during convolution, yielding faster inference times. However, the fixed nature of the algorithm may limit its applicability across various architectures and tasks, especially in scenarios where the network structure is highly dynamic or complex.

\cite{kivanc2024pyramid_training} introduced the pyramid training methodology, which reduces network complexity by progressively combining features across multiple sub-networks. While the approach achieves up to a 70\% reduction in model size, it faces challenges in effectively handling feature map combination, which can add additional computational costs. Additionally, the training of smaller sub-networks introduces overhead, potentially hindering its utility for real-time applications that require fast processing.

The Layer-Wise Complexity Reduction Method (LCRM) \citep{zhang2023lcrm} reduces the computational demands of CNNs by replacing standard convolutions with depthwise separable and pointwise convolutions. This strategy has shown promise in improving efficiency for models like AlexNet~\citep{krizhevsky2012imagenet} and VGG-9~\citep{simonyan2014very}. However, its application to more complex architectures is limited, and while it offers computational savings, it often results in a reduction in model performance, as the simplifications do not always preserve the discriminative power of the original network. Additionally, the method’s reliance on manual layer adjustments limits its flexibility in handling architectures outside of predefined models.

In contrast to tuning down overly complex model using the above methods, the approach proposed in this paper aims to build lightweight CNN models with much less complexity, in terms of model parameters and sizes, using lesser time and maintaining high accuracy by combining a few techniques in a creative manner. Moreover, the method shows great tendencies to self-regularize during training, reducing overfitting~\citep{srivastava2014dropout}, as the submodels in the model complement each other while increasing its data generalization ability. Furthermore, the steps are pretty straight forward and can be achieved in a well defined training pipeline.
The method further adds flexibility in that it allows you choose any model structure throughout the building process. Finally it is worth noting that any of the other optimizing methods could be applied to our model as this is more of another model building method or architecture than an optimization method for existing models, albeit, the aim is to train a relatively simpler and optimized model and potentially abort the need for applying further optimization methods.

\section{Methodology}
The methods describe in this paper consists of two main stages as shown in figure~\ref{fig:flowchart}: the first being the training of a dual-input-output model created and trained on the original and augmented version of the input dataset, the second involves the use of transfer learning method to fine tune the pre-trained model and and produce the final model with one input and one output.

\subsection{Datasets}
Three publicly available datasets were used in this experiment:
\begin{enumerate}
    \item MNIST handwriting dataset: The training dataset contained 60,000 images, split into 50,000 for training and 10,000 for validation, while the test dataset consisted of 10,000 images.
    \item fashion MNIST dataset: Same as the handwritten MNIST dataset.
    \item CIFAR-10 dataset: The training dataset contained 50,000 images, split into 40,000 for training and 10,000 for validation, while the test dataset consisted of 10,000 images.
\end{enumerate}

All datasets were obtained directly from TensorFlow's \texttt{tf.keras.datasets} library.

The datasets were converted to \texttt{tensorflow.data.Dataset} for better data pipeline streaming during training.

Note that for the two MNIST datasets, having shapes \texttt{[28x28]}, a third dimension was added to make them \texttt{[28, 28, 1]}, to satisfy CNN input requirements of \texttt{[height, width, channel]}, while the CIFAR-10 dataset was left unchanged (\texttt{[32x32x3]}).

\subsection{Data Preparation and Transformation}
According to the model shown in Figure~\ref{fig:concatenated_models}, the first sub-model was given the original data, while the second sub-model received the augmented version of the data using the \texttt{tf.keras.Sequential} pipeline.
The dataset was augmented using the following techniques:
\begin{itemize}
    \item \textbf{Random Rotation:} Applied with a factor of $\pm0.1$ radians.
    \item \textbf{Random Zoom:} Adjusted height and width by up to $\pm20\%$.
    \item \textbf{Random Brightness Adjustment:} Applied with a factor of 0 (minimal adjustment).
    \item \textbf{Random Translation:} Translated images by up to $\pm20\%$ of height and width.
\end{itemize}
The augmentation techniques used in this paper were not subjected to detailed technical considerations~\citep{shorten2019survey}; rather, they were applied to simulate common variations in images to a certain degree.

The dataset was processed using the TensorFlow \texttt{map} function to generate these paired inputs and outputs. The preprocessed training set for the second model was augmented dynamically, while the validation set remained unaltered. 
The final structure of the training and validation sets is shown below:

\begin{itemize}
    \item \textbf{Training Set:} Each sample was composed of the original image and its augmented counterpart, paired with their corresponding labels (actually same label). The mapping was as follows:  
    \[
    (X, y) \mapsto ((X, \text{augment}(X)), (y, y))
    \]
    \item \textbf{Validation Set:} Each sample included the original image paired with itself, ensuring no augmentation was applied during validation. The mapping was:  
    \[
    (X, y) \mapsto ((X, X), (y, y))
    \]
\end{itemize}

The augmentation process was implemented dynamically during training using the CPU, ensuring that the model received a varied input distribution in each epoch, a strategy supported by TensorFlow's design principles~\citep{abadi2016tensorflow}.

\subsection{Model Architecture}
The CNN model architecture, which was implemented using TensorFlow's \texttt{keras} framework, consists of concatenated sub-model layers and a dense layer.  

The model was initially created to contain two identical models with separate inputs and outputs. The inputs were fed two different sets of data: the first received the original dataset, and the second received the augmented version of the dataset. 
The concatenated models each had the following structure: 
\begin{enumerate}
    \item \textbf{Input Layer:}  
    The input shape is implicitly defined as \texttt{[28, 28, 1]} for the MNIST datasets and \texttt{[32, 32, 3]} for the CIFAR-10 dataset, corresponding to grayscale images of size 28x28x1 pixels for MNIST and color images of size 32x32x3 pixels for CIFAR-10.

    \item \textbf{First Convolutional Block:}  
    A 2D convolutional layer (\texttt{Conv2D}) with:
    \begin{itemize}
        \item \textbf{Number of Filters:} 10
        \item \textbf{Kernel Size:} $3 \times 3$ (used consistently across convolutional layers)
        \item \textbf{Activation Function:} ReLU
        \item \textbf{Weight Initialization:} He-initialization (\texttt{he\_normal})
        \item \textbf{Padding:} \texttt{valid} (no padding)
    \end{itemize}
    Followed by a max-pooling layer (\texttt{MaxPool2D}) to reduce spatial dimensions.

    \item \textbf{Second Convolutional Block:}  
    Similar to the first block but with 20 filters (double the previous block). This is followed by another max-pooling layer (\texttt{MaxPool2D}).

    \item \textbf{Flattening Layer:}  
    A \texttt{Flatten} layer to convert the 2D feature maps into a 1D vector, suitable for input into the fully connected layer.

    \item \textbf{Fully Connected Layer:}  
    A dense (\texttt{Dense}) layer with:
    \begin{itemize}
        \item \textbf{Number of Units:} 10 (corresponding to the number of classes in the classification task)
        \item \textbf{Activation Function:} Softmax (to output class probabilities)
    \end{itemize}
\end{enumerate}

The structure of the concatenated model is shown in Figure~\ref{fig:model1-model2} below.

\begin{figure}[!ht]
    \centering
    \includegraphics[width=0.2\linewidth, height=0.5\textheight]{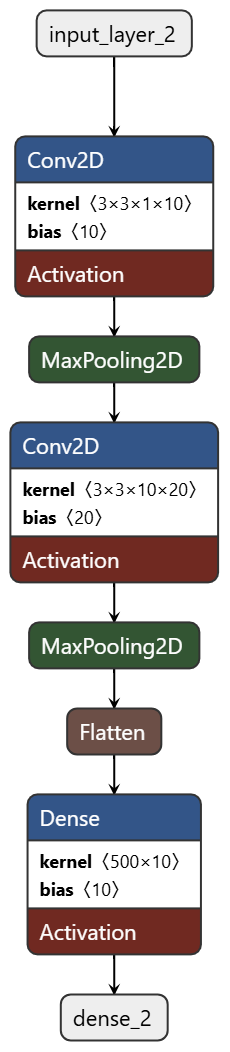}
    \caption{Structure of model2 (identical to  model1) before concatenation.}
    \label{fig:model1-model2}
\end{figure}

\begin{figure}[ht]
    \centering
    \includegraphics[width=0.5\linewidth]{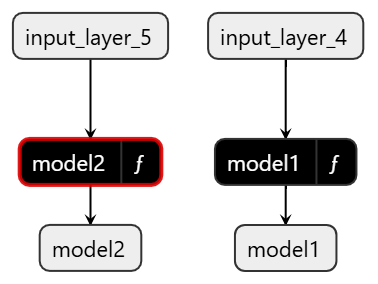}
    \caption{Concatenated model structure: here we have the inputs (\texttt{input\_layer\_5}, \texttt{input\_layer\_4}), the models (\texttt{model1 \textit{f}}, \texttt{model2 \textit{f}}) as described in Figure~\ref{fig:model1-model2}, and the outputs (\texttt{model1}, \texttt{model2}).}
    \label{fig:concatenated_models}
\end{figure}

Figure~\ref{fig:concatenated_models} shows the concatenated model structure.

\subsection{Model Compilation and Callbacks}
The model was compiled using the \texttt{Nadam} optimizer, with the evaluation metric set to \texttt{accuracy} and the loss function set to \texttt{sparse\_categorical\_crossentropy}. All other parameters were left at their default values. The choice of the Nadam optimizer, as supported by~\cite{depuru2019assessment}, is due to its ability to improve convergence speed and generalization in Convolutional Neural Networks (CNNs).

Two separate \texttt{ModelCheckpoint} callbacks were defined to monitor and save the best model weights: one for the first sub-model based on \texttt{val\_model1\_accuracy} and one for the second sub-model based on \texttt{val\_model2\_accuracy}. Both callbacks were set to save only the best-performing weights.

The \texttt{EarlyStopping} callback was employed to monitor the model's validation performance and halt training if no improvement was observed over a set number of epochs which is very effective in preventing model overfitting~\citep{bengio2012practical}. The early stopping mechanism was implemented as a functional callback with the following default parameter values:

\begin{itemize}
    \item \texttt{patience}: 5
    \item \texttt{monitor}: \texttt{'val\_accuracy'}
    \item \texttt{mode}: \texttt{'max'}
    \item \texttt{restore\_best\_weights}: \texttt{True}
    \item \texttt{min\_delta}: 0.001
\end{itemize}

These parameters were merely chosen because it was desired that the training has a patience of 5 and consider only an improvement of 0.001 increment as an improvement. There was nothing more technically considered than these.

\subsection{Model Training}
The model training was conducted on Kaggle using the NVIDIA Tesla P100-PCIE-16GB GPU resources and an Intel(R) Xeon(R) CPU @ 2.00GHz~\citep{kaggle_gpu_usage, kaggle_tesla_p100}. The GPU was utilized for training the models, while the CPU was used for data preprocessing tasks. All processes were consistent across all three datasets and their corresponding models; any modifications are explicitly highlighted in this paper.

\subsubsection{First Training Stage}
\begin{figure}[!h]
    \centering
    \includegraphics[width=1\linewidth]{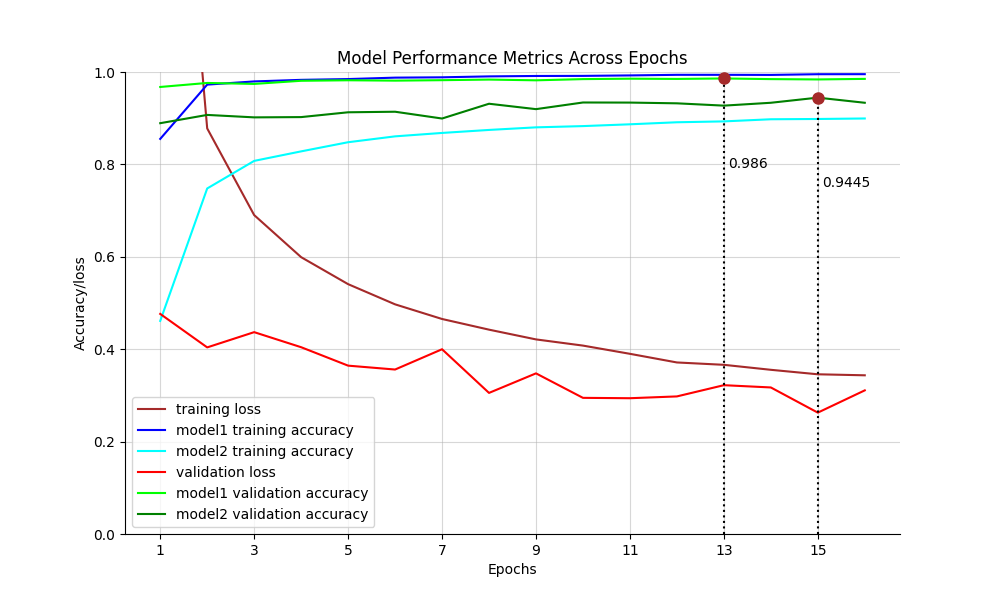}
    \caption{First training performance metrics of the two submodels for the handwritten MNIST dataset.}
    \label{fig:mnist_1_trainingl}
\end{figure}

\begin{figure}[H]
    \centering
    \includegraphics[width=1.0\linewidth]{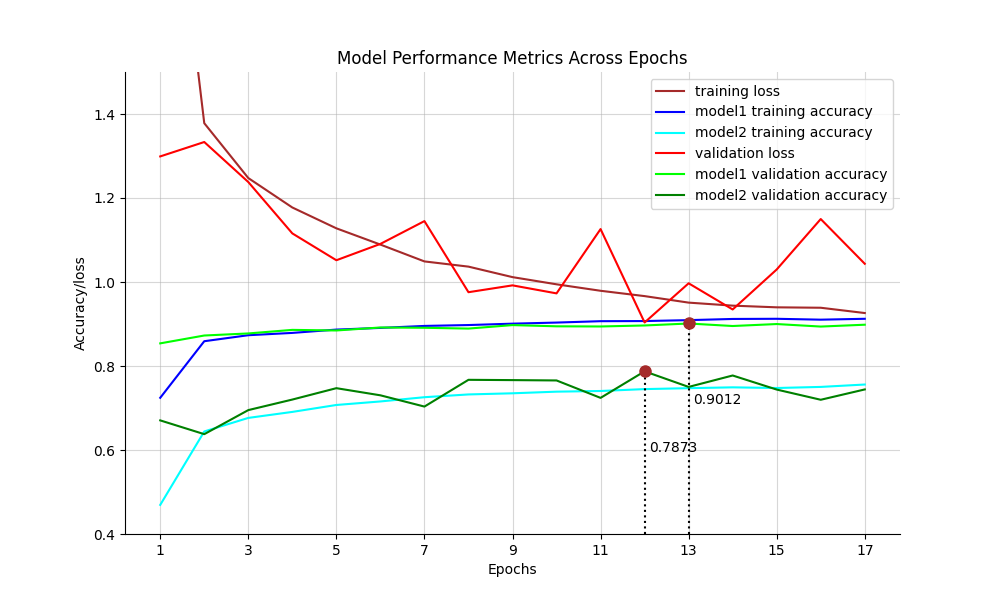}
    \caption{First training performance metrics of the two submodels for the fashion MNIST.}
    \label{fig:fashion_1_training}
\end{figure}

\begin{figure}[H]
    \centering
    \includegraphics[width=1.0\linewidth, height=0.39\textheight]{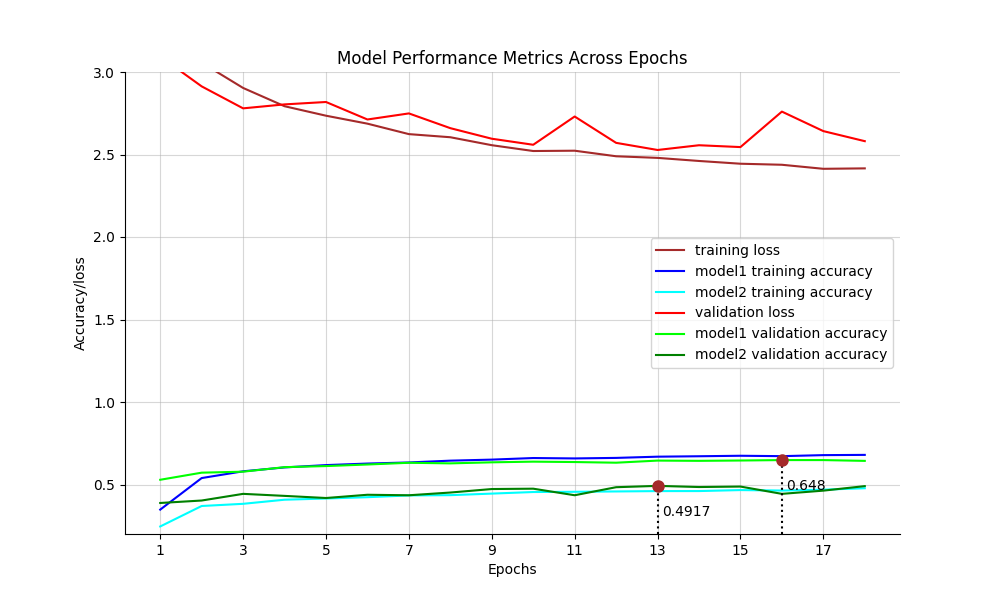}
    \caption{First training performance metrics of the two submodels for the cifar10 dataset.}
    \label{fig:cifar10_1_training}
\end{figure}

In the first training stage, the epoch was set to 20, but the models trained for lesser epochs before stopping. During the training, the performance of the two sub-models were monitored on the validation set. The models performance metrics are shown in Figure ~\ref{fig:mnist_1_trainingl} -~\ref{fig:cifar10_1_training}. For each model, as shown in the figure, the epoch at which the highest accuracy was achieved is marked by a dotted vertical line. The corresponding accuracy value is displayed next to the line, indicating the best model saved by the \texttt{ModelCheckpoint} callback. Note that in all the figures, submodel \texttt{model1} receives the original data and submodel, \texttt{model2}, receives the augmented data version.

\subsubsection{Creating the Final Model}

To create the final model, we first defined a new input layer. The previously trained model was loaded, and its layers were extracted. The original dense layers from each sub-model were removed and replaced with \texttt{Conv2D}. These convolutional layers were set with a kernel size of $(1, 1)$, a stride of $(1, 1)$, and \texttt{'same'} padding, ensuring that the convolution operation would mimic the fully connected behavior of the dense layer while maintaining spatial locality.

The number of filters in the \texttt{Conv2D} layers was set to match the number of units (10) in the original dense layers, which allows the convolutional layers to capture similar learned features. Additionally, the weights of the dense layers were used to initialize the corresponding convolutional layers. This was done by reshaping the dense layer's weight matrix into the appropriate format for the convolutional kernels, ensuring that the final model retained the learned features from the initial training. This describes the techniques involved in the conversion of a dense layer to a CNN layer as discussed by~\cite{long2015fully_convolutional}.

After concatenating the outputs of the two convolutional layers, the result was flattened and passed through a dense layer with 32 neurons, 50\% dropout rate was added to prevents overfitting ~\citep{srivastava2014dropout}, and a ReLU activation function. This dense layer was designed to capture the combined features from both sub-models. Finally, a softmax output layer was added to produce predictions for the 10 classes.
It is worth noting that from this stage onward, further trainings made used of the original dataset and not the augmented version. This has no technical reasons than that the aim of the augmentation was to have a form of a submodel that could capture more variability that might not have been in the original dataset. Hence after the first training phase it was never used anymore, of course future work can look into this in details.

\begin{figure}[!h]
    \centering
    \includegraphics[width=0.3\linewidth, height=0.5\textheight]{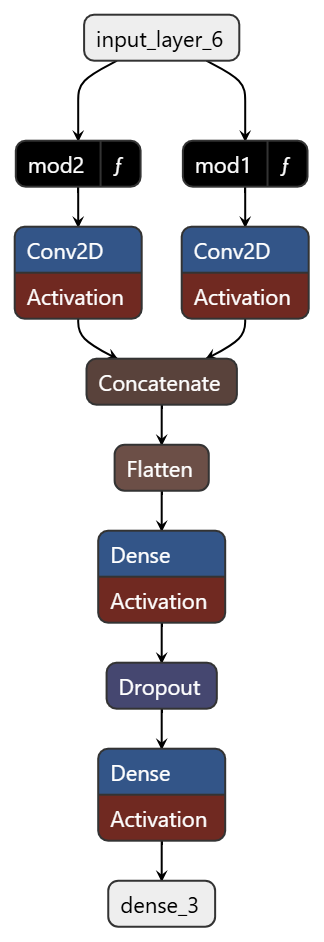}
    \caption{Final model architecture: \texttt{mod1 \textit{f}} and \texttt{mod2 \textit{f}} represent \texttt{model1} and \texttt{model2} without their dense layers; the next \texttt{Conv2D} layers on either side represent the converted dense layers.}
    \label{fig:enter-label}
\end{figure}

\subsubsection{Second (Final) Training Stage}
The second training stage which is also the final stage consists of the following stages: 
\begin{itemize}
    \item \textbf{Stage 1:}
    In the first stage, the model's layers were frozen except for the last two layers (the newly added layers). The model was compiled using the same parameters as before. Early stopping was applied during training to prevent overfitting, and the model was trained for 20 epochs with early stopping based on validation accuracy. The training metrics are shown in Figure~\ref{fig:mnist_2_training}-~\ref{fig:cifar10_2_training}.
    
    \begin{figure}
        \centering
        \includegraphics[width=1\linewidth]{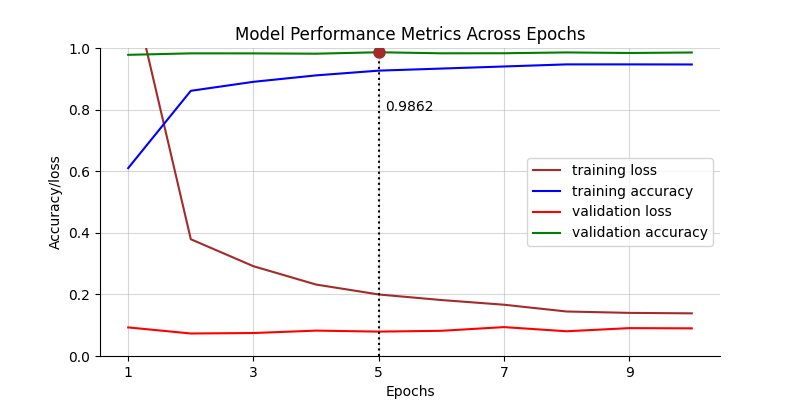}
        \caption{Second training performance metrics of the model for the MNIST handwritten dataset.}
        \label{fig:mnist_2_training}
    \end{figure}
    
    \begin{figure}
        \centering
        \includegraphics[width=1\linewidth]{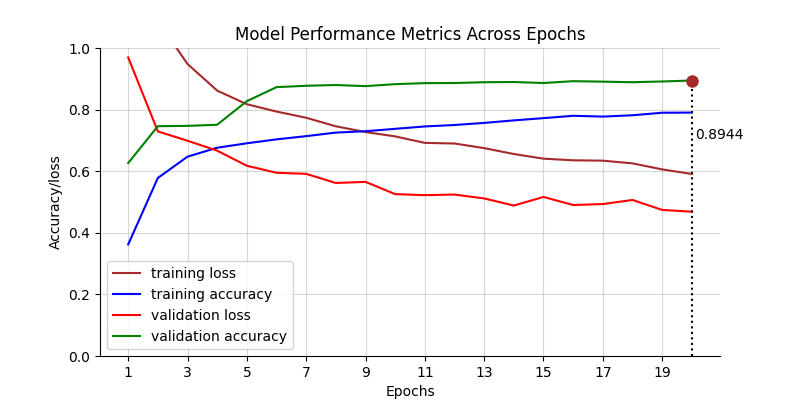}
        \caption{Second training performance metrics of the model for the fashion MNIST dataset. It is worth noting that the model shows a trend of continual improvement up until the 20th epoch, but increasing the epochs did not show any further improvement; hence, I rolled back to 20 epochs for consistency in reporting}
        \label{fig:fashion_2_training}
    \end{figure}
    
    \begin{figure}
        \centering
        \includegraphics[width=1\linewidth]{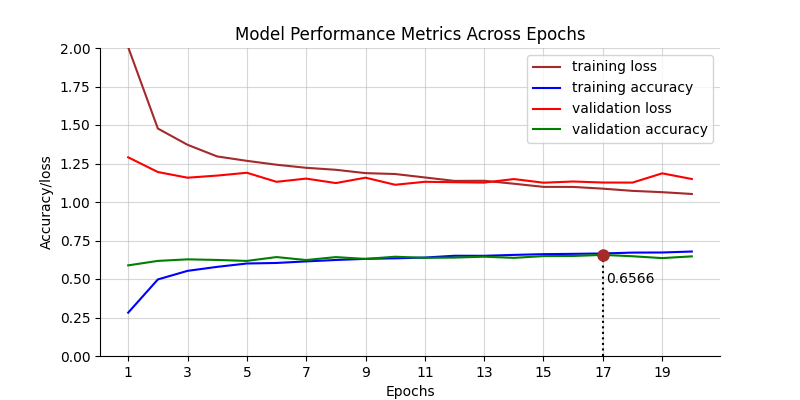}
        \caption{Second training performance metrics of the model for the Cifar10 dataset.}
        \label{fig:cifar10_2_training}
    \end{figure}
        
    \item \textbf{Stage 2:}
    After completing the first stage, all the layers of the model were unfrozen, and the model was recompiled using the Stochastic Gradient Descent (SGD) optimizer. The optimizer was configured with the following parameters:
    \begin{itemize}
        \item Learning rate: 0.001
        \item Momentum: 0.9
        \item Weight decay: $1 \times 10^{-4}$
        \item Nesterov momentum: True
    \end{itemize}
    The use of SGD was in order to provide finer control over weight updates and reduce the risk of overfitting as~\cite{kumar2022fine_tune_sgd} highlight; SGD is particularly effective in fine-tuning deep neural networks due to its ability to adapt weight adjustments with greater precision, to ensure robust generalization and efficient memory utilization.
    The model was set to train for up to 50 epochs, but early stopping was applied to terminate training if the validation accuracy did not improve for 5 consecutive epochs. However, training stopped before the 20th epoch for all three models due to the early stopping callback. The training metrics are shown in Figure~\ref{fig:mnist_3_training} -~\ref{fig:cifar10_3_training}.
    
    \begin{figure}[!h]
        \centering
        \includegraphics[width=1\linewidth]{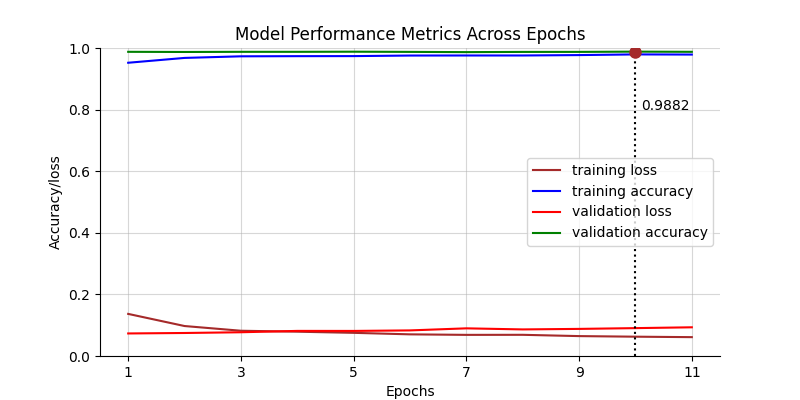}
        \caption{Third training performance metrics of the two submodels for the MNIST handwritten dataset.}
        \label{fig:mnist_3_training}
    \end{figure}
    
    \begin{figure}[!h]
        \centering
        \includegraphics[width=1\linewidth]{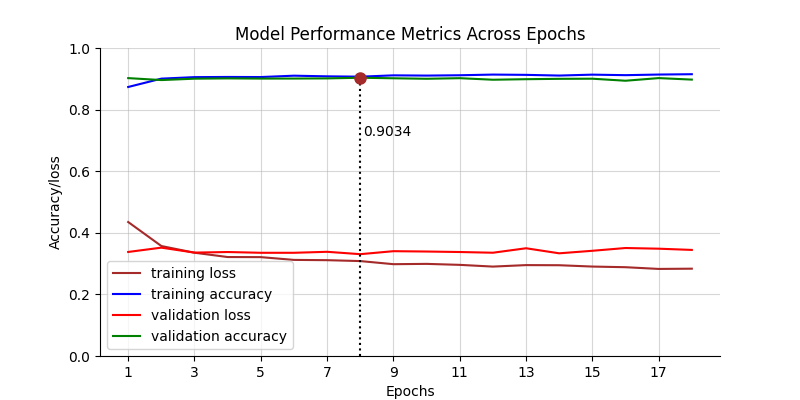}
        \caption{Third training performance metrics of the two submodels for the fashion MNIST handwritten dataset.}
        \label{fig:fashion_3_training}
    \end{figure}
    
    \begin{figure}[!h]
        \centering
        \includegraphics[width=1\linewidth]{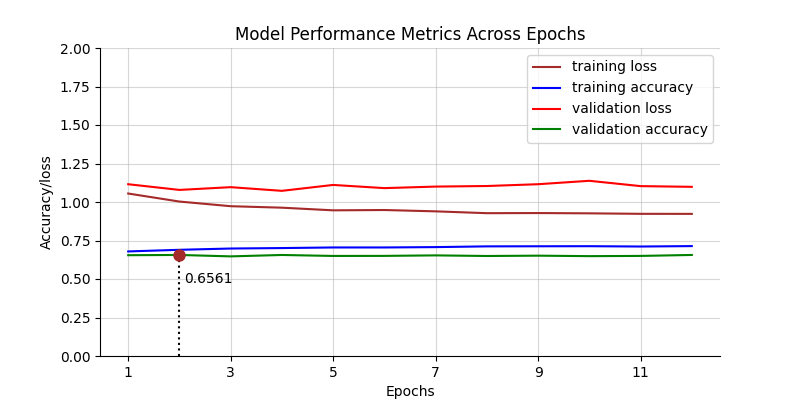}
        \caption{Third training performance metrics of the two submodels for the cifar10 handwritten dataset.}
        \label{fig:cifar10_3_training}
    \end{figure}
    
    \item \textbf{Stage 3:}
    In the final step, K-fold cross-validation was performed with six splits. This choice was primarily based on the MNIST and fashion MNIST datasets, which each contained 60,000 images, allowing for a 10,000-image validation set per fold. However, the same consideration was not applied to the CIFAR-10 dataset, resulting in a validation set with fewer than 10,000 images. Training for each fold utilized the SGD optimizer with consistent parameters. The training was conducted for a maximum of 10 epochs, but early stopping was triggered before reaching the full 10 epochs. The longest training session lasted four epochs before stopping. K-fold cross-validation was chosen for its reliability in assessing model performance and ensuring robust generalization~\citep{Kohavi1995cross_validation}. For each k-fold, the best model was saved by the checkpoint callback, but at the end, the final accuracy on the test set for each of the models including the overall model was approximately the same to the third decimal.
\end{itemize}

\section{Results and Discussion}

The results of these experiments are not directly compared to other models, instead it is demonstrated in different aspects how effective the model is. Seeing that the model is very light-weighted, the focus is in what the model had achieved. Moreover, to the best of my knowledge, even LeNet~\citep{lecun1998gradient} which was designed specifically for handwritten digit MNIST and was later applied to fashion MNIST had about 60,000 parameters and reached the same accuracy as the model with less than 14,900 parameters proposed in this paper.

\subsection{Training Performance}

As stated earlier, a consistent architecture and training process were maintained across all three datasets, with the exception of adjusting the model input shape: \texttt{[32x32x3]} for CIFAR-10 and \texttt{[28x28x1]} for the two MNIST datasets.

The model performance for the datasets is summarized in Table~\ref{tab:results}. Across the datasets, the lightweight model demonstrated strong accuracy, with a consistent trend in both training and test set accuracy.

\begin{table*}
    \centering
    \resizebox{\textwidth}{!}{
    \begin{tabular}{|c|c|c|}
    \hline
    \textbf{Dataset} & \textbf{Approx. Final Training Accuracy (\%)} & \textbf{Approx. Final Test Set Accuracy (\%)} \\
    \hline
    handwritten digit MNIST & 100 & 99 \\
    \hline
    fashion MNIST & 91 & 89 \\
    \hline
    CIFAR-10 & 73 & 65 \\
    \hline
    \end{tabular}
    }
    \caption{Summary of model performance across different datasets}
    \label{tab:results}
\end{table*}

From table~\ref{tab:results}, it can be seen that that the model performed much better on the handwritten digit MNIST and the fashion MNIST datasets; The consistent results between these two datasets suggest that the model is well balanced in simplicity and accuracy for both datasets. However, the accuracy dropped much more with the cifar10 dataset, a more complex dataset with color images and higher variability, indicating that the model might be too simple for the dataset and could benefit from slight modifications to improve performance as discussed in~\nameref{sec:future_work}.

\begin{figure}[!h]
    \centering  
    \includegraphics[width=0.8\linewidth]{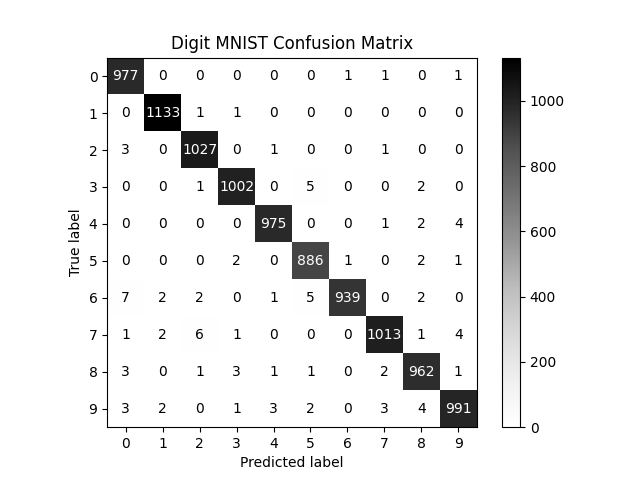}
    \caption{Confusion matrix of the MNIST-trained model evaluated on the MNIST test set.}
    \label{fig:mnist_cm}
\end{figure}

\begin{figure}[!h]
    \centering
    \includegraphics[width=0.7\linewidth]{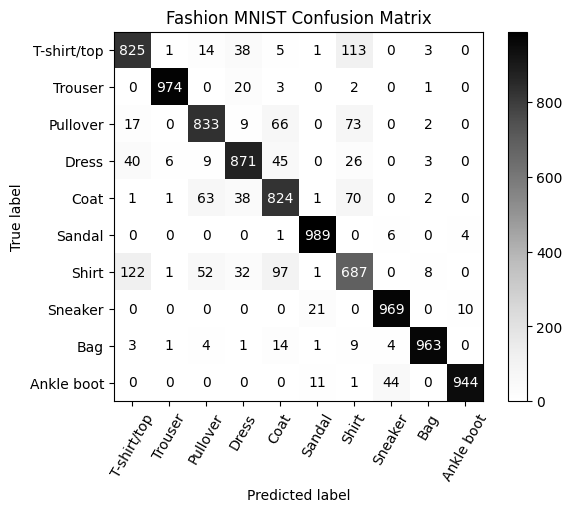}
    \caption{Confusion matrix of the fashion MNIST-trained model evaluated on the fashion MNIST test set.}
    \label{fig:fashion_mnist_cm}
\end{figure}

\begin{figure}[!h]
    \centering
    \includegraphics[width=0.7\linewidth]{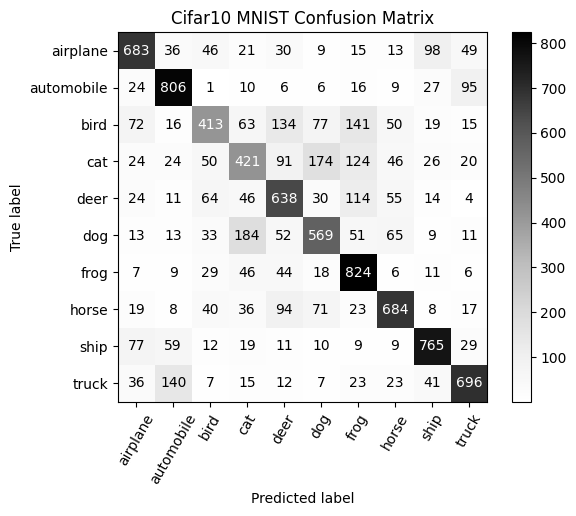}
    \caption{Confusion matrix of the cifar10-trained model evaluated on the Cifar10 test set.}
    \label{fig:cifar10_cm}
\end{figure}

The confusion matrix for the datasets, evaluated on the test sets, are shown in Figure~\ref{fig:mnist_cm} to~\ref{fig:cifar10_cm}.

Additionally, the classification report for each dataset, evaluated on their corresponding trained models, is shown in tables~\ref{tab:mnist_classification_report} to~\ref{tab:cifar10_classification_report}.
\begin{table}[ht]
    \centering
    \begin{tabular}{lcccc}
    \hline
    \textbf{Class} & \textbf{Precision} & \textbf{Recall} & \textbf{F1-Score} & \textbf{Support} \\ \hline
    0 & 0.98 & 1.00 & 0.99 & 980 \\
    1 & 0.99 & 1.00 & 1.00 & 1135 \\
    2 & 0.99 & 1.00 & 0.99 & 1032 \\
    3 & 0.99 & 0.99 & 0.99 & 1010 \\
    4 & 0.99 & 0.99 & 0.99 & 982 \\
    5 & 0.99 & 0.99 & 0.99 & 892 \\
    6 & 1.00 & 0.98 & 0.99 & 958 \\
    7 & 0.99 & 0.99 & 0.99 & 1028 \\
    8 & 0.99 & 0.99 & 0.99 & 974 \\
    9 & 0.99 & 0.98 & 0.99 & 1009 \\ \hline
    \textbf{Accuracy} & \multicolumn{4}{c}{0.99 (10,000 samples)} \\ \hline
    \textbf{Macro Avg} & 0.99 & 0.99 & 0.99 & 10,000 \\
    \textbf{Weighted Avg} & 0.99 & 0.99 & 0.99 & 10,000 \\ \hline
    \end{tabular}
    \caption{Digit MNIST Classification Report}
    \label{tab:mnist_classification_report}
\end{table}

\begin{table}[ht]
    \centering
    \begin{tabular}{lcccc}
    \hline
    \textbf{Class} & \textbf{Precision} & \textbf{Recall} & \textbf{F1-Score} & \textbf{Support} \\ \hline
    0 & 0.82 & 0.82 & 0.82 & 1000 \\
    1 & 0.99 & 0.97 & 0.98 & 1000 \\
    2 & 0.85 & 0.83 & 0.84 & 1000 \\
    3 & 0.86 & 0.87 & 0.87 & 1000 \\
    4 & 0.78 & 0.82 & 0.80 & 1000 \\
    5 & 0.96 & 0.99 & 0.98 & 1000 \\
    6 & 0.70 & 0.69 & 0.69 & 1000 \\
    7 & 0.95 & 0.97 & 0.96 & 1000 \\
    8 & 0.98 & 0.96 & 0.97 & 1000 \\
    9 & 0.99 & 0.94 & 0.96 & 1000 \\ \hline
    \textbf{Accuracy} & \multicolumn{4}{c}{0.89 (10,000 samples)} \\ \hline
    \textbf{Macro Avg} & 0.89 & 0.89 & 0.89 & 10,000 \\
    \textbf{Weighted Avg} & 0.89 & 0.89 & 0.89 & 10,000 \\ \hline
    \end{tabular}
    \caption{fashion MNIST Classification Report}
    \label{tab:fashion_classification_report}
\end{table}

\begin{table}[ht]
    \centering
    \begin{tabular}{lcccc}
    \hline
    \textbf{Class} & \textbf{Precision} & \textbf{Recall} & \textbf{F1-Score} & \textbf{Support} \\ \hline
    0 & 0.70 & 0.68 & 0.69 & 1000 \\
    1 & 0.72 & 0.81 & 0.76 & 1000 \\
    2 & 0.59 & 0.41 & 0.49 & 1000 \\
    3 & 0.49 & 0.42 & 0.45 & 1000 \\
    4 & 0.57 & 0.64 & 0.60 & 1000 \\
    5 & 0.59 & 0.57 & 0.58 & 1000 \\
    6 & 0.61 & 0.82 & 0.70 & 1000 \\
    7 & 0.71 & 0.68 & 0.70 & 1000 \\
    8 & 0.75 & 0.77 & 0.76 & 1000 \\
    9 & 0.74 & 0.70 & 0.72 & 1000 \\ \hline
    \textbf{Accuracy} & \multicolumn{4}{c}{0.65 (10,000 samples)} \\ \hline
    \textbf{Macro Avg} & 0.65 & 0.65 & 0.64 & 10,000 \\
    \textbf{Weighted Avg} & 0.65 & 0.65 & 0.64 & 10,000 \\ \hline
    \end{tabular}
    \caption{Cifar10 Classification Report}
    \label{tab:cifar10_classification_report}
\end{table}

\subsection{Model Lightweightness}
The primary aim of this paper is to create a lightweight CNN model that maintains very high accuracy fast inference time \citep{breck2020mlnet}. The lightweight nature of the model is demonstrated by measuring its latency, throughput, and size. Notably, the final models were not subjected to any post-training quantization or modifications to reduce their size or improve runtime performance.

As mentioned earlier, the training of these models were carried out on the kaggle platform using the NVIDIA Tesla P100-PCIE-16GB GPU resources and an Intel(R) Xeon(R) CPU @ 2.00GHz. Hence the following evaluations would be based on these hardware.

\subsubsection{Model Latency}
Table~\ref{tab:latency_comparison} shows the latency of each model evaluated over a single instance and averaged over 100 repetitions. the standard deviation is also shown for both of the GPU and the CPU.

\begin{table*}[ht]
    \centering
    \resizebox{\textwidth}{!}{
    \begin{tabular}{|l|c|c|c|c|}
    \hline
    \textbf{Dataset} & \textbf{CPU Latency (ms)} & \textbf{CPU Std (ms)} & \textbf{GPU Latency (ms)} & \textbf{GPU Std (ms)} \\ \hline
    MNIST            & 13.22                     & 1.15                  & 11.22                     & 0.20                  \\ \hline
    fashion MNIST    & 13.04                     & 0.47                  & 11.23                     & 0.21                  \\ \hline
    CIFAR-10         & 13.65                     & 0.58                  & 11.23                     & 0.19                  \\ \hline
    \end{tabular}}
    \caption{Latency and standard deviation (ms) for CPU (Intel(R) Xeon(R) CPU @ 2.00GHz) and GPU (NVIDIA Tesla P100-PCIE-16GB) across different datasets. GPU consistently outperforms CPU with lower latency and standard deviation.}
    \label{tab:latency_comparison}
\end{table*}

\subsubsection{Model Throughput}
\begin{figure*}    
    \centering
    \begin{minipage}{0.49\linewidth}
        \centering
        \includegraphics[width=\linewidth]{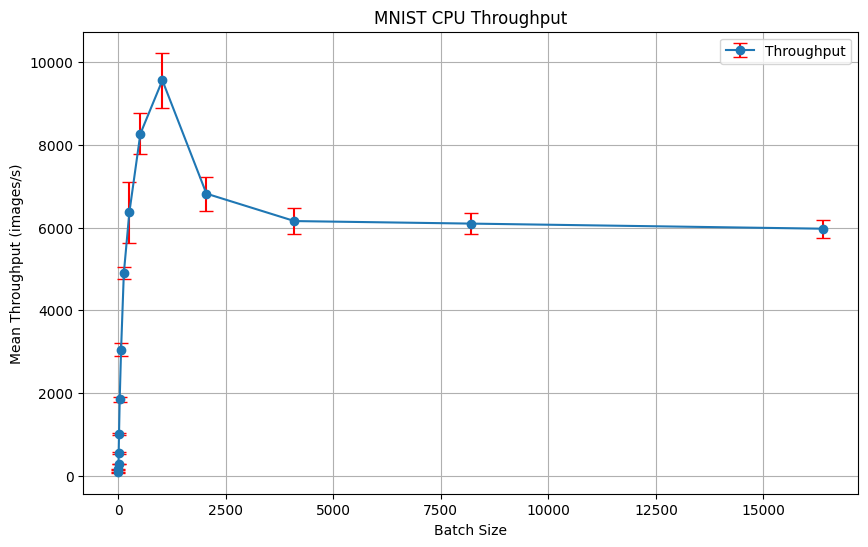}
        \caption{MNIST CPU Throughput}
        \label{fig:MNIST CPU Throughput}
    \end{minipage}
    \hfill
    \begin{minipage}{0.49\linewidth}
        \centering
        \includegraphics[width=\linewidth]{MNIST_CPU_Throughput.png}
        \caption{MNIST GPU Throughput}
        \label{fig:MNIST GPU Throughput}
    \end{minipage}
    
    \vspace{0.5cm}
    
    \begin{minipage}{0.49\linewidth}
        \centering
        \includegraphics[width=\linewidth]{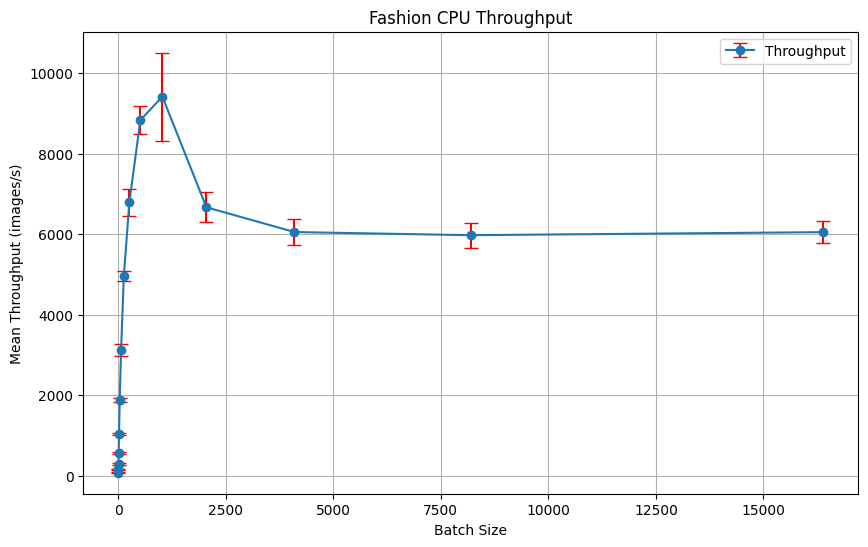}
        \caption{fashion MNIST CPU Throughput}
        \label{fig:Fashion CPU Throughput}
    \end{minipage}
    \hfill
    \begin{minipage}{0.49\linewidth}
        \centering
        \includegraphics[width=\linewidth]{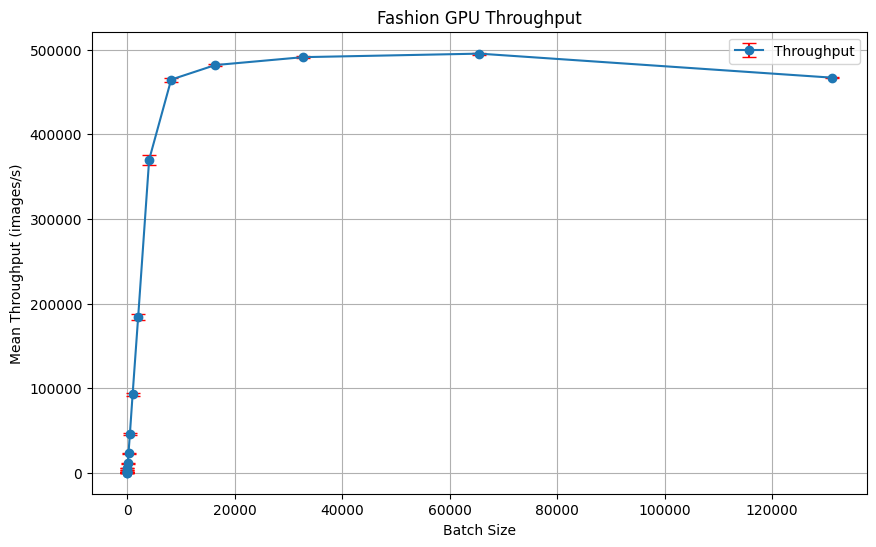}
        \caption{fashion MNIST GPU Throughput}
        \label{fig:Fashion GPU Throughput}
    \end{minipage}
    
    \vspace{0.5cm}
    
    \begin{minipage}{0.49\linewidth}
        \centering
        \includegraphics[width=\linewidth]{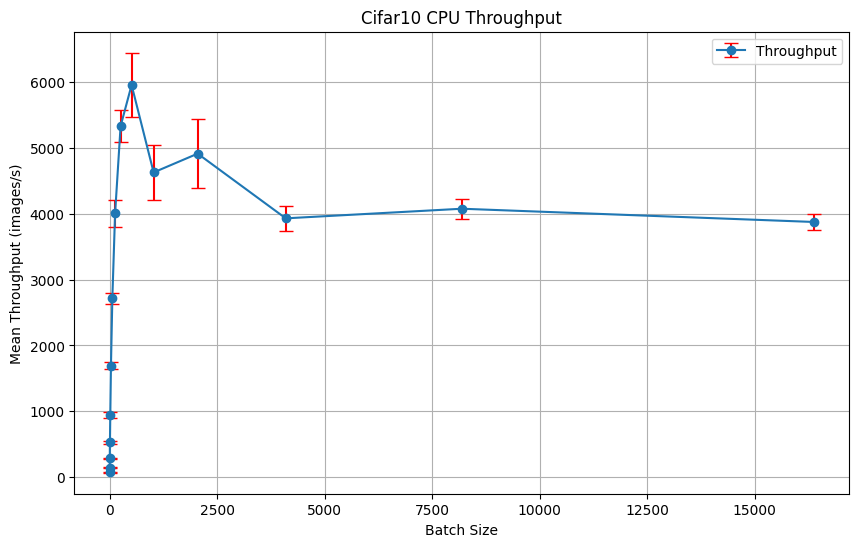}
        \caption{CIFAR-10 CPU Throughput}
        \label{fig:Cifar10 CPU Throughput}
    \end{minipage}
    \hfill
    \begin{minipage}{0.49\linewidth}
        \centering
        \includegraphics[width=\linewidth]{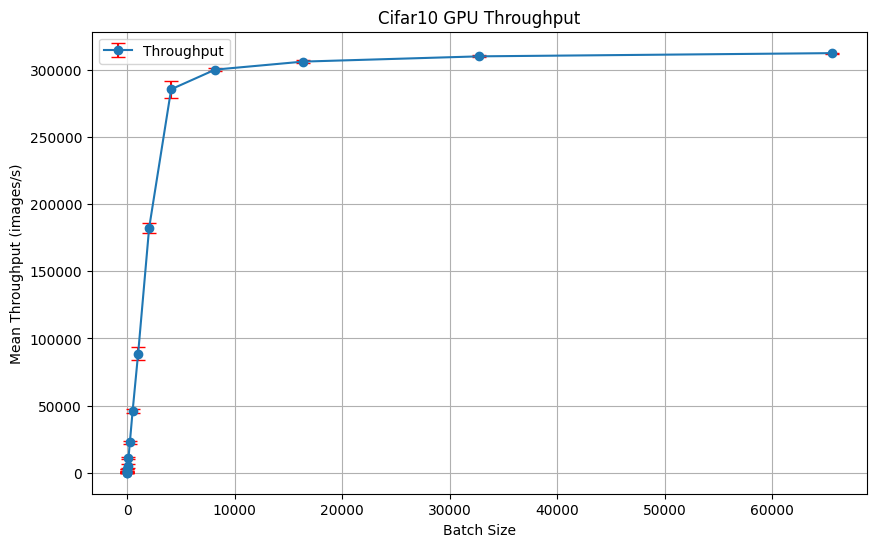}
        \caption{CIFAR-10 GPU Throughput}
        \label{fig:Cifar10 GPU Throughput}
    \end{minipage}
\end{figure*}

Figure~\ref{fig:MNIST CPU Throughput} to \ref{fig:Cifar10 GPU Throughput} shows the throughput of each model evaluated over different batch sizes, which increase exponentially in powers of 2 (i.e., $2^n$, where $n = 0, 1, 2, \dots$), and averaged over 100 repetitions for each batch size. The standard deviation is represented with red error bars. The graphs show that throughput values vary with the batch sizes. It is worth noting that for the CPU, the batch size was increased up to $n = 14$, while for the GPU, the batch size was continually increased until a TensorFlow resource exhaustion error occurred. The error was captured, and the process was terminated at that particular batch size. Therefore, the optimal batch size for the GPU is likely between the highest throughput value and the batch size where the error occurred, or between the highest throughput value and either adjacent batch size. It is worth noting that latency is better evaluated at the optimal batch size \cite{zhao2020inference_time}. This batch size could only be determined through grid search, incrementing the batch size by one. Furthermore, the Kaggle platform produced varying results even when grid search was used, sometimes throwing errors at certain batch sizes and other times not. As a result, batch sizes increasing exponentially in powers of two were chosen for consistency.

\subsubsection{Model Size}
Model sizes and parameter counts are shown in table~\ref{tab:model_size} 

\begin{table}[ht]
    \centering
    \begin{tabular}{|l|l|l|}
    \hline
    \textbf{Dataset}      & \textbf{Model Size} & \textbf{Trainable Params} \\ \hline
    MNIST                 & 0.17 MB             & 14,862 (58.05 KB)         \\ \hline
    fashion MNIST         & 0.17 MB             & 14,862 (58.05 KB)         \\ \hline
    CIFAR-10              & 0.21 MB             & 19,622 (76.65 KB)         \\ \hline
    \end{tabular}
    \caption{Model Sizes and Trainable Parameters for MNIST, fashion MNIST, and CIFAR-10 Models.}
    \label{tab:model_size}
\end{table}

\subsection{Discussion}
The results from the three datasets demonstrate the effectiveness of the proposed lightweight model architecture and training methodology in tackling diverse image classification tasks. Each dataset was used to train the same model architecture independently, employing identical processes, and the outcomes were evaluated separately.

The consistent use of the same architecture across all datasets without any structural modifications underscores the flexibility and robustness of the model. For smaller, grayscale datasets such as handwritten digit MNIST and fashion MNIST, the lightweight architecture performed exceptionally well reaching a a the value of 99\% and 89\% for all of the metrics respectively as shown on table~\ref{tab:mnist_classification_report} and~\ref{tab:fashion_classification_report}. However, CIFAR-10 had a much lower metrics scores of 65\% except fo its F1-Score which has a value of 64\% as in table ~\ref{tab:cifar10_classification_report}.

When we look at the confusion matrix of all three datasets, figure~\ref{fig:mnist_cm} to~\ref{fig:cifar10_cm}, gotten from their final test sets, it is easily observed that some classes contributed more to the errors made by the model than other; these includes class "shirt" and "T-shirt/top" for the fashion MNIST being confused for each other, and class "bird", "cat", "deer" etc in the cifar10 dataset.
This deviation on the Cifar10 dataset can be attributed to the dataset, having a more complex structure of colored images. Hence a higher probability that the model used in this paper is overly simple for the task. 

For all three first model training, it can be seen from the graphs, figure~\ref{fig:mnist_1_trainingl} to~\ref{fig:cifar10_1_training}, that the two submodels in the model generally improved along the training epochs with their errors dropping and their accuracies going upward. The same trend can be seen for the second phase of the training, figure~\ref{fig:mnist_2_training} to ~\ref{fig:cifar10_2_training}, and the the third phase, figure~\ref{fig:mnist_3_training} to~\ref{fig:fashion_3_training}. However, the cifar10 model, figure~\ref{fig:cifar10_3_training}, ceased to improve its validation accuracy and started to overfit instead.

It can be seen that little energy was given to addressing overfitting in the whole project because the method proves to be self-regularizing. For example during the first training phase, the highly performing submodel, fed on the original dataset, overfits, while the one with the augmented version does not. During concatenation, this two complemented each other, leading to a balanced model that does not overfit. However towards the end of the training phase, the optimizer parameters, SGD in this case, is fine tuned to have a more weight update control mechanism.

The model latency shown in table~\ref{tab:latency_comparison} shows just how fast the model is, relative to the hardware, at making inference in a production environment both in a CPU and a GPU environment. these value is relatively constant across all models and dataset.
Finally, the model sizes and parameters reported in table~\ref{tab:model_size} shows that the model has very small parameters compared to a lot of larger models created for small tasks such as the MNIST dataset. Despite this small size, the model was able to reach a benchmark accuracy of 99\% for the MNIST dataset.

Unlike existing methods, our approach enables small model initialization and inherently addresses overfitting. It achieves robust performance by separately leveraging the original and augmented datasets, enhancing generalization while preserving critical patterns. However, during transfer learning phases involving layer freezing and unfreezing, careful selection of optimizers is critical for maintaining performance stability.

It is important to emphasize that the aim of this work was not to compare performance across datasets or to evaluate the generalization of a single model to multiple datasets. Instead, the focus lies in demonstrating the efficacy of the lightweight model and the associated training methodology. Future research could explore further optimizations, such as enhanced data augmentation, advanced regularization techniques, or architecture refinements, to improve performance on more complex datasets.

\section{Conclusion}

This paper proposes a unique combination of methods for building lightweight Convolutional Neural Networks (CNNs) that achieve high accuracy while maintaining reduced complexity. The approach leverages two key strategies:

\begin{enumerate}
    \item \textbf{Dual-Input-Output Models:} The architecture incorporates two submodels in one model; one trained on the original dataset and the other trained on an augmented version. Concatenating their outputs later on enables the model to capture features from both data variations, enhancing robustness and reducing overfitting on smaller datasets.

    \item \textbf{Transfer Learning with Progressive Unfreezing:} After independent training of the submodels withing the modther model, their weights are combined into a single model and few new layers are added. This unified model undergoes transfer learning, with layers gradually unfrozen for fine-tuning. This allows the new layers to learn and to adapt and utilizes pre-learned features, this finally leads to a final model with high robustness.
\end{enumerate}

The proposed method offers several advantages which includes:
\begin{itemize}
    \item \textbf{Reduced Complexity:} The the method used through the used use of two simple identical submodels allows for the creation of a less complex architecture instead of one gigantic model, specifically seen in the fact that hyper-parameters like number of filters were initialized to small numbers. 
    \item \textbf{Improved Robustness:} Training on both original and augmented data mitigates overfitting, particularly on smaller datasets, capturing data variability.
    \item \textbf{Competitive Accuracy:} As demonstrated in the results, table~\ref{tab:results}, the approach achieves high accuracy across benchmark datasets even with fewer parameters.
\end{itemize}

\subsection*{Future Work}\label{sec:future_work}
For the improvement of the results and strategies used in this paper, future work can take care of the following limitations or proposals:
\begin{itemize}
    \item Explore the applicability of this method to various image recognition tasks beyond MNIST, fashion MNIST and the cifar10 dataset.
    \item Investigate the impact of different hyperparameter settings (e.g., number of filters, learning rate) on the performance of the model, but note that the aim is to keep the model simple, hence the initialization should also be relatively simple.
    \item Formulate rules to follow in initializing hyper-parameters such as number of filters, number of convolutional layers, number of dense layers as well as the number of neurons, so as to keep the overall process simple during model creation and initialization and not against the essence of the paper which seeks to keep models light-weighted.
    \item Explore a much better data augmentation technique and the unfreezing order that would foster better performance of the model~\citep{shorten2019survey, howard2018universal}.
    \item Analyze a better model stoppage techniques other than the early stopping used in this project.
    \item Analyze the effectiveness of the method on larger and more complex datasets to assess its scalability.
    \item Compare the proposed approach with other state-of-the-art techniques for complexity reduction in CNNs, such as pruning and quantization.\item Come up with a base formula or rule of thumb for deciding what model would not be complex beyond its given task.
    \item Propose a method for rebuilding existing models into a fraction of their size for the same accuracy using the proposed method in this paper.
    \item Use a more powerful platform for training and evaluation for a more accurate result.
    \item Extension of this approach to tasks beyond image classification should be explored.
\end{itemize}

Addressing these aspects can further refine and solidify the proposed method as a valuable tool for developing highly efficient and lightweight CNNs, particularly for applications with limited computational resources or where faster inferences is of high demand.

A lightweight model need not always be as small as the one demonstrated in this paper. However, this work underscores the transformative potential of reconstructing existing models with only a fraction of their original size and parameters while retaining comparable performance. For instance, consider a model with 1 billion parameters—rebuilding such a model with just a million parameters (or a comparable value), achieving nearly identical or better results, then such a rebuilded-model would indeed be considered lightweight compared to the parent model. This highlights the feasibility of drastically reducing model complexity without compromising accuracy, paving the way for more efficient and accessible deep learning solutions.

\bibliography{reference}

\end{document}